\documentclass[runningheads]{llncs}

\usepackage{eccv}

\usepackage{eccvabbrv}

\usepackage{graphicx}
\usepackage{booktabs}
\usepackage{bbm}
\usepackage{tabu}

\usepackage[accsupp]{axessibility}  

\usepackage[pagebackref,breaklinks,colorlinks,citecolor=eccvblue]{hyperref}

\usepackage{orcidlink}

\newcommand{\tokens}{\mathcal{T}}

\begin{document}

\title{Enhancing Semantic Fidelity in Text-to-Image Synthesis: Attention Regulation
in Diffusion Models} 

\titlerunning{Attention Regulation in Diffusion Models}

\author{Yang Zhang\inst{1}, Teoh Tze Tzun\inst{1}, Lim Wei Hern\inst{1}, Tiviatis Sim\inst{1}, Kenji Kawaguchi\inst{1}}

\authorrunning{Y.Zhang et al.}

\institute{National University of Singapore \\
\email{\{yangzhang, teoh.tze.tzun, limweihern, tiviatis\}@u.nus.edu, kenji@comp.nus.edu.sg}
}
\maketitle

\begin{abstract}
Recent advancements in diffusion models have notably improved the perceptual quality of generated images in text-to-image synthesis tasks. However, diffusion models often struggle to produce images that accurately reflect the intended semantics of the associated text prompts. We examine cross-attention layers in diffusion models and observe a propensity for these layers to disproportionately focus on certain tokens during the generation process, thereby undermining semantic fidelity. To address the issue of dominant attention, we introduce \emph{attention regulation}, a computation-efficient on-the-fly optimization approach at inference time to align attention maps with the input text prompt. Notably, our method requires no additional training or fine-tuning and serves as a plug-in module on a model. Hence, the generation capacity of the original model is fully preserved. We compare our approach with alternative approaches across various datasets, evaluation metrics, and diffusion models. Experiment results show that our method consistently outperforms other baselines, yielding images that more faithfully reflect the desired concepts with reduced computation overhead. Code is available at \url{https://github.com/YaNgZhAnG-V5/attention_regulation}.
\end{abstract}

\section{Introduction}
Diffusion models introduce a significant paradigm shift in the field of generative models \cite{ho2020denoising, song2020score, yang2022diffusion}, with their application becoming increasingly widespread. Their adoption of diffusion models is largely attributed to their capability to generate detailed, high-resolution, and diverse outputs across a broad spectrum of domains. Moreover, diffusion models excel in leveraging conditioned inputs for conditional generation. This adaptability to various forms of conditions, whether textual or visual, further extends the application of diffusion models beyond mere image generation, encompassing tasks such as text-to-video synthesis\cite{wu2023tune}, super-resolution\cite{lin2023magic3d}, image-to-image translation\cite{saharia2022palette}, and image inpainting \cite{lugmayr2022repaint}. 

\begin{figure}
    \centering
    \includegraphics[width=\textwidth]{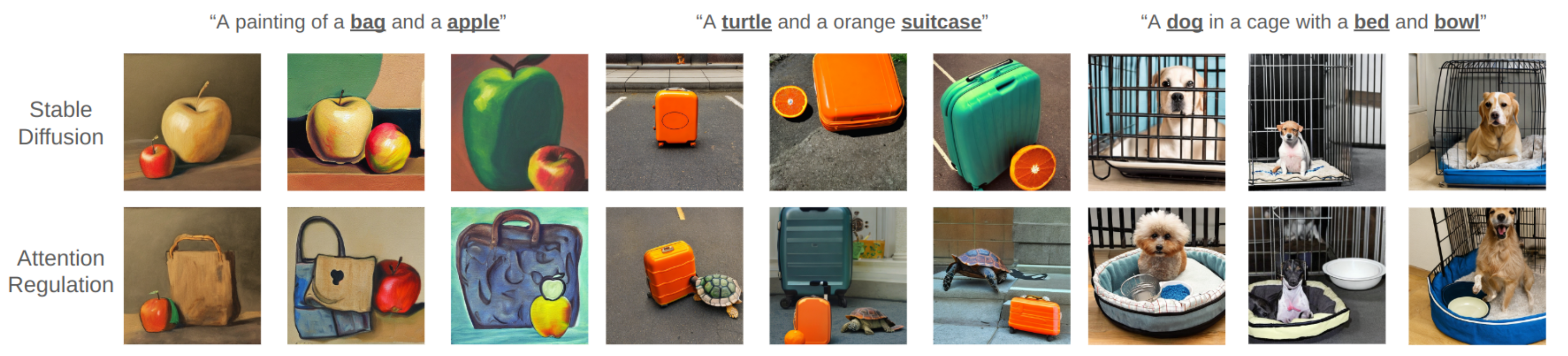}
    \caption{Attention regulation effectively improves semantics alignment with prompts by modifying the cross-attention maps at inference time without fine-tuning the model. Moreover, attention regulation requires only additional information on target tokens and achieves inference time comparable to that of the original model. Attention regulation serves as a plug-in module and can be disabled anytime to use the original model. }
    \label{fig:first_figure}
\end{figure}
Although diffusion models are adept at producing images of high perceptual quality, they face challenges in following specific conditions for image generation. This limitation is particularly pronounced in text-to-image (T2I) synthesis as compared to tasks where generation relies on more descriptive conditions, such as segmentation masks or partial images used in inpainting. Previous studies have observed that diffusion models can ignore some tokens in the input prompt during the generation process, a problem known as ``catastrophic neglect''\cite{chefer2023attendandexcite} and ``missing objects'' \cite{feng2022training}. Additionally, these models may overly focus on certain aspects of the prompt, resulting in generation results that are excessively similar to data encountered during the training phase. Prior works proposed to improve the stability of diffusion models by learning to use additional input as conditions, such as sketches or poses as visual cues, or additional instructions that the model should follow. However, these approaches may require additional training or fine-tuning the model. In addition, these methods demand more inputs, potentially restricting their usability in scenarios where acquiring additional conditional information is challenging.

In this work, we introduce attention regulation, a method that modifies attention maps within cross-attention layers during the reverse diffusion process to better align attention maps with desired properties. We formulate the desired regulation outcome through a constrained optimization problem. The optimization aims to enhance the attention of all target tokens while ensuring that modifications to the attention maps remain minimal and essential. Our attention regulation technique obviates the need for additional training or fine-tuning, enabling straightforward integration with existing trained models. Furthermore, it selectively targets a subset of cross-attention layers for optimization, thereby reducing computational demands and minimizing inference time. Experimental outcomes demonstrate the superior efficacy of our attention regulation approach, significantly improving the semantic coherence of generated images with comparably less computational overhead during inference against baseline methods. Examples of Attention regulation is shown in Figure~\ref{fig:first_figure}.

\textbf{Contribution: }
a) We propose an on-the-fly attention edit method on T2I diffusion models to improve their textual following ability. We formulate the attention edit problem as a constrained optimization problem on attention maps and show how it can be solved by gradient-based optimization. b) We propose an evaluation metric based on a detection model while evaluating our proposal with existing evaluation metrics. Evaluation across various diffusion models and datasets demonstrates the effectiveness of our method. 
\section{Related Works}
Diffusion models simulate the process where noise gradually obscures source data until it becomes entirely noisy \cite{sohl2015deep}. The goal is to learn the reverse process, allowing a model to recover data from noisy inputs. Diffusion models can either predict less noisy data at each step or deduce the noise, then denoise data using Langevin dynamics. \cite{ho2020denoising, saharia2022photorealistic}. Earlier diffusion models reconstruct images from noise directly, a computation-intensive process that limits reconstruction speed. Instead, Rombach et al.\cite{rombach2022latent_diffusion} processing in a lower-dimensional space, making latent diffusion models significantly faster and enabling training on extensive datasets like LAION. \cite{schuhmann2022laion}. Prediction of noise or previous states in the reverse process usually uses a U-Net \cite{ronneberger2015unet}. To enable conditional generation for diffusion models, cross-attention modules \cite{vaswani2017attention} are embedded into the U-Net so that generation takes the condition into account \cite{rombach2022latent_diffusion}. Other guidance techniques are proposed to improve the conditional generation performance\cite{ho2021classifier,song2020score,dhariwal2021diffusion}. 

To gain more control over diffusion models, various approaches propose to edit trained models through fine-tuning \cite{ruiz2023dreambooth}\cite{Kawar_2023_CVPR}. Custom diffusion \cite{kumari2023multi} proposes to fine-tune a diffusion model to include customized objects and achieve image composition including customized objects. Concept Erasing \cite{gandikota2023erasing} and Concept Ablation \cite{kumari2023conceptablation} work by removing target concepts given by users.
Besides fine-tuning models to enhance control, alternative methods modify the diffusion process without fine-tuning to guide the model generation. Null Text Inversion \cite{gal2022image} optimizes a text embedding to elicit specific behaviors from the diffusion model, utilizing this embedding to steer the generation process.
Prompt-to-Prompt \cite{hertz2022prompttoprompt} edits the content of a generated image by interchanging attention maps of different prompts. 
Composable Diffusion \cite{liu2023compositional} assembles multiple diffusion models, utilizing each of them to model an image component.  
Syntax-Guided Generation (SynGen) \cite{rassin2024linguistic} conducts syntactic analysis of prompts to identify entities and their relationships, then utilizes a loss function to encourage the cross-attention maps to agree with the linguistic binding.
Dense Diffusion \cite{kim2023dense} edits an image by modifying the attention map of a given target object using a segmentation mask, which is markedly different from our approach as our method does not require a predefined segmentation mask.
Attend-and-Excite \cite{chefer2023attendandexcite} addresses catastrophic neglect, a tendency to neglect information from prompts during image generation. The difference between Attend-and-Excite and our attention regulation approach is that they optimize the latent variable based on a loss function defined over attention maps, while ours directly regulates attention maps. 

\section{Method}
\subsection{Preliminaries}\label{sec:preliminaries}
\textbf{Diffusion models. }
Diffusion models constitute a class of generative models that simulate the physical process of diffusion. In a diffusion process, Gaussian noise is incrementally introduced to the original data across multiple steps, transforming the data samples into pure noise. The objective of diffusion models is to learn the reverse diffusion process that converts noise back into data conforming to the target data distribution. This reverse process can be effectively modeled by learning to predict the noise at a specific diffusion step, denoted as $\hat{\epsilon}_\theta(x_t, t)$. The loss function is thus defined as
\begin{equation}
    \mathcal{L} = \sum_{t=1}^{T} \mathbb{E}_{x_0, \epsilon\sim\mathcal{N}(\mu, \sigma^2), t} \left[ \left\| \epsilon - \hat{\epsilon}_\theta(x_t, t) \right\|^2 \right], 
\end{equation}
where \(x_t\) represents a noisy version of the data \(x\), and \(t\) is uniformly sampled from \(\{1, \ldots, T\}\). To improve the sample efficiency of diffusion models, one can transform the data into a low-dimensional hidden space using a Variational Autoencoder (VAE). Given an encoding model \(\mathcal{E}(\cdot)\), the hidden representation \(z\) of the data \(x\) is obtained as \(z = \mathcal{E}(x)\). In addition, a diffusion model can learn conditional distributions \(P(x|c)\) using a conditional denoising model. A more comprehensive loss function, incorporating initial conditions and latent representation, is
\begin{equation}
        \mathcal{L} = \sum_{t=1}^{T} \mathbb{E}_{\mathcal{E}(x_0), c, \epsilon\sim\mathcal{N}(\mu, \sigma^2), t} \left[ \left\| \epsilon - \hat{\epsilon}_\theta(z_t, t, c) \right\|^2 \right].
\end{equation}

\textbf{Cross-attention layers in diffusion models. }
The previous section discussed the conditional generation ability of diffusion models. Conditional information is incorporated into diffusion models through cross-attention layers. A cross-attention layer typically has many attention heads. The functional representation of an attention head, \(f(z_t,c)\), which integrates the hidden representation $z_t\in\mathbb{R}^{M\times d_z}$ and the condition \(c\) contains N tokens, is defined as
\begin{equation}
    f(z_t, c) = \text{Attention}(Q, K, V) = \text{softmax}\left(\frac{QK^T}{\sqrt{d_k}}\right) \cdot V,
\end{equation}
where \(Q=W_Q\cdot z_t\), \(K=W_K\cdot\tau_{\theta}(c)\), and \(V=W_V\cdot\tau_{\theta}(c)\). In this formulation, the model $\tau_{\theta}$ transforms the condition \(c\) into a latent conditional representation $\tau_{\theta}(c)\in\mathbb{R}^{N \times d_c}$, then projects $\tau_{\theta}(c)$ and $z_t$ into key $K\in\mathbb{R}^{N \times d}$, query $Q\in\mathbb{R}^{M \times d}$, and value $V\in\mathbb{R}^{N \times d}$ through weights $W_Q\in\mathbb{R}^{d\times d_z}$, $W_K\in\mathbb{R}^{d\times d_c}$, and $W_V\in\mathbb{R}^{d\times d_c}$. Lastly, $K$, $Q$, and $V$ are processed by the attention mechanism. We can extract an attention map $A$ as
\begin{equation}
    A = \text{softmax}\left(\frac{QK^T}{\sqrt{d_k}}\right)\in\mathbb{R}^{M \times N}.
\end{equation}
Attention map $A$ provides the correlation in terms of attention scores between the hidden representation and the condition. An attention map $A$ can be further processed by unraveling the first dimension to be $N$ two-dimensional maps, where each map shows the attention of one text token on the image. 

\subsection{Semantic Violation by Attention Mismatch}\label{sec:semantic_violation}
\begin{figure}[t]
    \centering
    \includegraphics[width=0.95\textwidth]{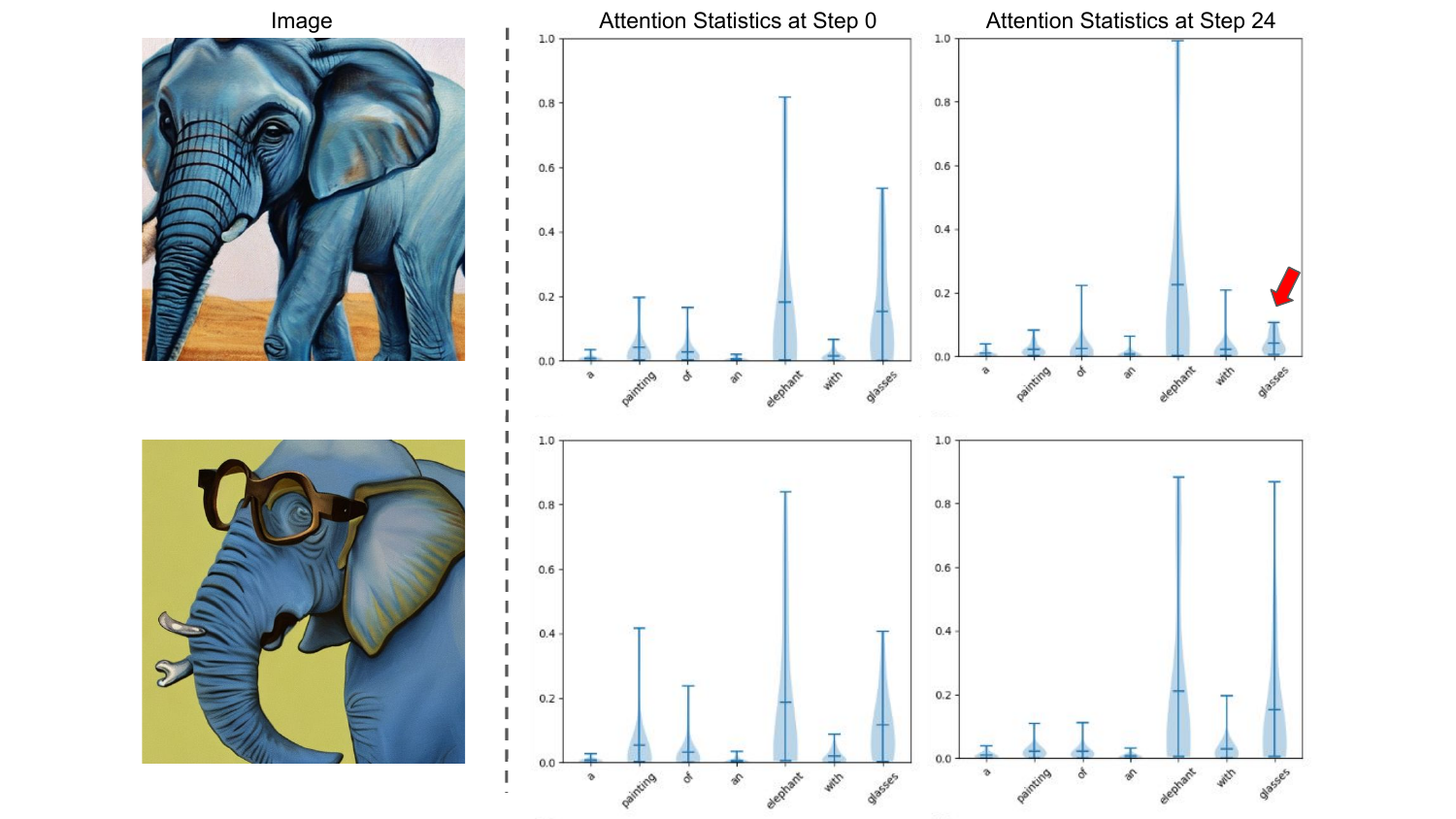}
    \caption{Illustration of attention dominance. The violin plots display the attention statistics for one cross-attention layer across two image samples, both prompted by "A painting of an elephant with glasses." At the initial diffusion step $0$ (middle column), the attention patterns are similar for both samples. By step $24$ (third column), a significant divergence is evident. For the successful sample (bottom row), the attention allocated to "elephant" and "glasses" is approximately equal, suggesting a balanced representation. In contrast, for the sample that fails to include glasses (top row), attention disproportionately favors the token "elephant," marginalizing other relevant tokens (red arrow). More results are in the Appendix~\ref{appx:more_attention_stats}
    .}
    \label{fig:show_issue}
\end{figure}
In this section, we investigate why diffusion models fail to adhere to the semantics of a given prompt. As outlined in Section~\ref{sec:preliminaries}, cross-attention layers are responsible for integrating conditional information such as prompt embeddings. Therefore, our analysis focuses on the functioning of these cross-attention layers during the reverse diffusion process. 
Figure~\ref{fig:show_issue} illustrates the attention statistics for two samples generated in response to the same prompt, "A painting of an elephant with glasses", albeit with differing initial noises. The implementation and visualization details of this experiment are provided in Appendix~\ref{appx:more_attention_stats}. Notably, one sample successfully includes glasses in the image, while the other sample fails to generate glasses. 
The attention statistics for both samples initially exhibit a similar pattern during the early stages of the reverse diffusion process. However, at diffusion step $24$, the attention statistics for the unsuccessful sample (the one lacking glasses) reveal a predominance of the "elephant" token in attention values. Given that this sample ultimately fails to include glasses, we conjecture that this disproportionate focus detracts from the representation of other relevant tokens, thereby diminishing the semantic integrity of the generated image. In addition, dominance attention usually appears during the generation process instead of at the initial states of the reverse diffusion. This pattern of dominant attention is observed across multiple cross-attention layers and throughout various diffusion steps. Furthermore, such instances of dominating attention, particularly in samples with semantically incorrect outcomes, are a common occurrence. For generated samples with correct textual semantics, the attention is more evenly distributed across relevant tokens. Additional examples are provided in Appendix~\ref{appx:more_attention_stats} as a more comprehensive demonstration. Conditional diffusion models usually apply guidance, we also provide additional results in Appendix~\ref{appx:guidance_scale} that using a larger guidance scale is not sufficient to solve the semantic violation issue.

To mitigate this effect of dominant attention during the reverse diffusion process, we propose attention regulation in the subsequent section. Attention regulation employs an intuitive way to improve image semantics based on our observation on attention maps: \emph{we should mitigate dominating attention and promote attention of all relevant tokens}.

\subsection{Attention Edit as Constrained Optimization}
\textbf{Optimization objective. }In Section~\ref{sec:semantic_violation}, we show that attention mismatch during the reverse diffusion process causes the generated images to deviate from the intended semantics. To improve the semantic fidelity of generated images, we introduce attention regulation, a method that applies on-the-fly adjustments to attention maps at inference time. We formulate this attention edit process as a constrained optimization problem based on the notation in Section~\ref{sec:preliminaries}. In our attention regulation setting, we require a set of target token indexes $\mathcal{T}=\{t_1, ..., t_n\}$ as additional input. We want to ensure sufficient attention on target tokens during the reverse diffusion process, such that the final image contains objects representing target tokens. Formally, for a given set of original attention maps $A\in\mathbb{R}^{M\times N}$ prior to adjustment, and an error function $E(\cdot, \cdot)$ that quantifies the deviation of target attention maps from desired characteristics, the optimally edited attention map $A^*$ is defined as follows:
\begin{align}
    A^* =& \quad\arg\min_{A^{\prime}} E(A^{\prime}, \mathcal{T}) \\
    \text{subject to} & \quad ||A^{\prime} - A||_2\leq\delta,
\end{align}
where \(\delta\) represents a threshold for the allowable deviation from the original attention maps. This constrained optimization problem can be converted to an unconstrained optimization problem by introducing a Lagrange multiplier $\beta > 0$ for the inequality constraint. The resulting optimization problem, in terms of the Lagrangian $f(A, \beta)$, becomes:
\begin{equation}
    A^* = \arg\min_{A^{\prime}, \beta} f(A^{\prime}, \beta, \mathcal{T}) = E(A^{\prime}, \mathcal{T}) + \beta\cdot (||A^{\prime} - A||_2 - \delta).
\end{equation}
We further set $\beta$ as a non-negative hyperparameter and omit all constants in the equation, this yields:
\begin{equation}
    A^* = \arg\min_{A^{\prime}} L(A^{\prime}, \mathcal{T}) = E(A^{\prime}, \mathcal{T}) + \beta\cdot ||A^{\prime} - A||_2.
\end{equation}
Our optimization aims to mitigate attention dominance and encourage the attention maps of all target tokens to have sufficiently high attention values. Therefore, we define the error function $E(\cdot)$ as: 
\begin{equation}
    E(A^{\prime}, \mathcal{T}) = \frac{1}{|\tokens|}\sum_{t \in \tokens} (\phi(A^{\prime}_t, 0.9) - 0.9)^2 + \alpha\cdot\frac{1}{|\tokens|}\sum_{t \in \tokens} \left(\sum_{a\in A_t^{\prime}}a - \mu\cdot M \right)^2,
\end{equation}
where $A_t^{\prime}\in\mathbb{R}^M$ is a 2D attention map of a target token, $\phi(A_t^{\prime}, 0.9)$ extracts the $90^{th}$-quantile of all the value in $A_t^{\prime}$, and $\sum_{a\in A_t^{\prime}}a$ calculates the sum of all elements in $A_t^{\prime}$. The first term in $E(A^{\prime}, \mathcal{T})$ aims to ensure that high attention regions in the attention maps of each target token reach a specified threshold, ideally so that the $90^{th}$ quantile of the target attention map equals $0.9$. The intuition behind the second term of $E(A^{\prime}, \mathcal{T})$ is to ensure that there is an equal $\mu$ proportion of high attention region in the attention map of each target token. This formulation of $E(A^{\prime}, \mathcal{T})$ results in a differentiable loss function $L(A^{\prime}, \mathcal{T})$, allowing for gradient-based optimization to find \(A^*\). The subsequent sections will elaborate on the methodology to parameterize $A^{\prime}$ and optimization details.

\textbf{Parametrize attention maps. }
Given a query matrix $Q$ and a key matrix $K$ extracted from a cross-attention layer, we parametrize edited attention maps as
\begin{equation}
    A^{\prime} = M_A(S) = softmax\left(\frac{QK^T + S}{\sqrt{d}}\right),
\end{equation}
where $S$ is an additive adjustment to the query-key-product $QK^T$ from the cross-attention layer. 
This parameterization allows for effective modification of the attention scores while preserving their normalization property, wherein the attention scores across the token dimension sum up to one.

\begin{figure}[t]
    \centering
    \includegraphics[width=0.95\textwidth]{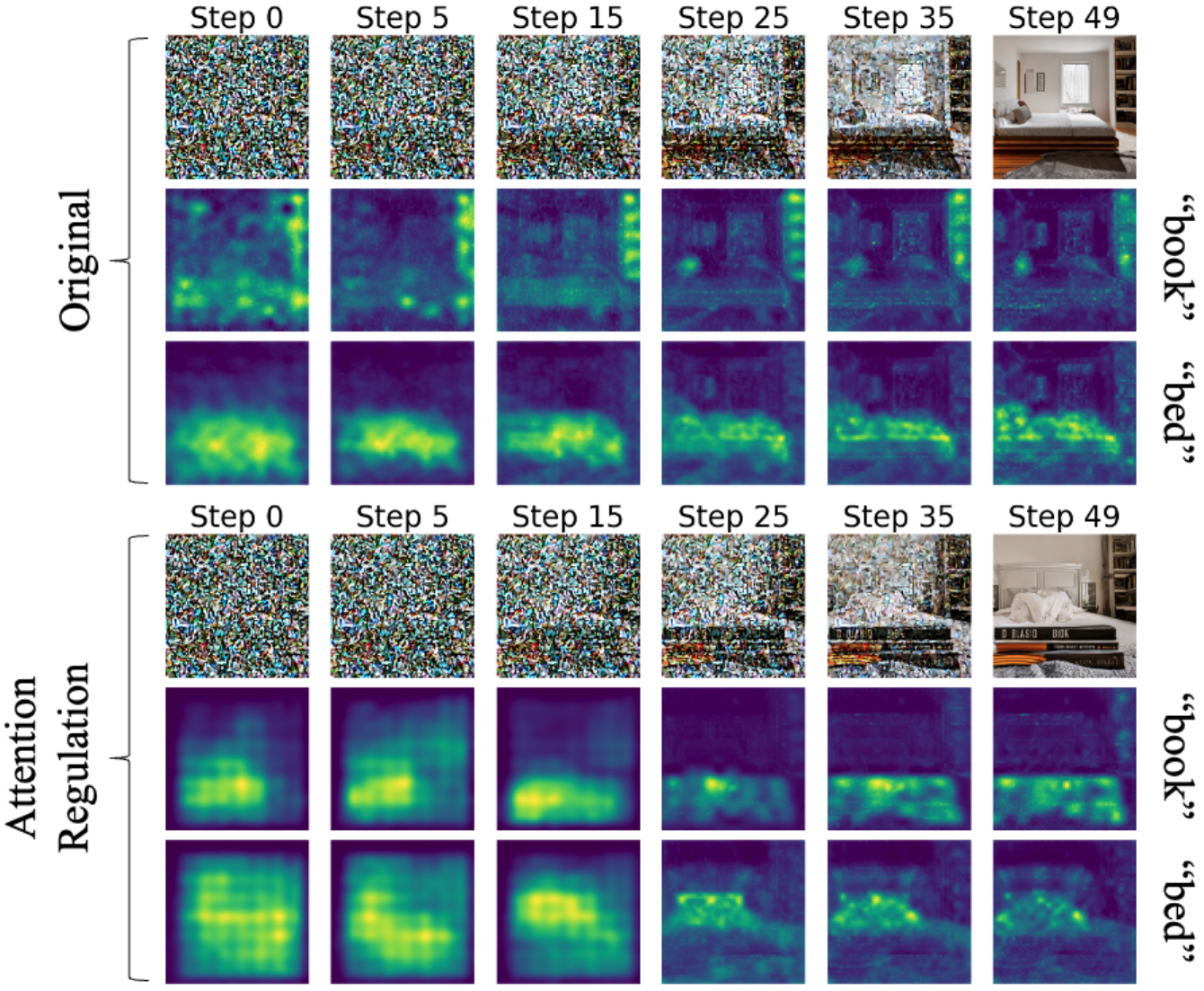}
    \caption{Visualization of the optimization outcome. The given prompt is "A bedroom with a book on the bed". By creating regions with high attention values for the target tokens while maintaining the consistency of the attention maps across diffusion steps, the desired targets are successfully generated.}
    \label{fig:opt_visualization}
\end{figure}

To minimize artifacts during editing and expedite the optimization process, we aim for $S$ to be smooth and parameterized by another variable with fewer trainable parameters. 
Consequently, we further parameterized $S$ using a weight matrix $\theta\in\mathbb{R}^{r\times r}$ with $r = \frac{w}{2\sigma}$ and $w^2 = M$. $S$ is then defined as
\begin{equation}
    S = M_S(\theta, \sigma) = \sum_{p = 1}^{r} \sum_{q = 1}^{r} \theta_{p, q}\cdot G(2\sigma p, 2\sigma q, \sigma),
\end{equation}
where matrix $G(x_0, y_0, \sigma)\in\mathbb{R}^{w \times w}$ represents a 2D smooth Gaussian kernel, expressed by
\begin{equation}
    G(x_0, y_0, \sigma) =  \exp\left(-\frac{(i - y_0)^2 + (j - x_0)^2}{2\sigma^2} \right)_{1 \leq i, j \leq w}.
\end{equation}
Parameter $\sigma$ is chosen such that $2\sigma$ divides $w$. 
This approach shifts the focus of optimization from the entire attention map to merely learning a weight matrix $\theta$ for the smooth additive variable $S$, described by
\begin{equation}
    A^{\prime}\leftarrow M_A(M_S(\theta - \eta\cdot\nabla_{\theta}L)),
\end{equation}
where $\eta$ denotes the learning rate. With this parameterization of attention maps, we have only $\frac{M}{4\sigma^2}$ learnable parameters instead of $M$ learnable parameters. This strategy ensures a targeted and efficient adjustment of attention maps, thereby enhancing semantic fidelity in generated images with minimal computational overhead. Figure~\ref{fig:opt_visualization} shows a visualization of the target attention maps after optimization at specific diffusion steps.

\textbf{Reduce distortion in generation. }While our optimization objective aims to minimize edits on attention maps, the extent of its regulation effect is strongly influenced by the hyperparameter $\beta$. In practice, $\beta$ is usually suboptimal, leading to overediting of the attention maps. Such overediting in attention maps can introduce distortions into the generated images. We identify two primary causes of distortion during the generation process with attention regulation. First, optimization may alter different spatial regions at each diffusion time step. These spatially inconsistent attention maps across diffusion steps can contribute to distortion in generated images. Second, substantial edits during the later stages of the reverse diffusion process can adversely affect the generation, as later reverse diffusion steps are responsible for adding fine-grained visual details. To address these two identified issues, we introduce the following attention map edit scheme:
\begin{align}
    A_{EMA} &\leftarrow \kappa\cdot A_{EMA} + (1 - \kappa)\cdot A^*, \\
    A &\leftarrow \lambda^t\cdot \mathbbm{1}_{t< t_{\text{thres}}}(t)\cdot A_{EMA}.
\end{align}
Rather than directly applying the optimized result, we calculate an Exponential Moving Average (EMA) of the optimized result, \(A_{EMA}\), as an additional consistency enforcement. Moreover, we apply a decay to the edit variable as reverse diffusion progresses, gradually decreasing the impact of the edit, aligning with approaches in prior works \cite{kim2023dense} that also gradually reduce the amount of edit. Lastly, we stop the edit beyond a certain diffusion time step threshold.

\textbf{Efficient attention regulation. }We selectively apply attention regulation only on a subset of all cross-attention layers. As T2I diffusion models usually apply a U-Net structure for noise prediction, we choose cross-attention layers in the last down-sampling layers and the first up-sampling layers in the U-Net for editing. This targeted approach allows us to concentrate our editing efforts on layers that have a significant impact on the model's ability to incorporate and refine semantic details, optimizing the balance between fidelity to the text prompt and efficiency. Section~\ref{sec:ablation_study} shows that this design choice achieves a good trade-off between efficiency and performance.

\section{Experiments}

\subsection{Experiment Setup}
\textbf{Baselines: }We restrict our comparison against four other training-free methods proposed to improve semantic fidelity: Composable diffusion, Syntax Generation, Dense Diffusion and Attend-And-Excite. 

\textbf{Evaluation Metrics: } We apply five metrics for evaluation. For Semantic Alignment evaluation, we use CLIP score (denoted as CLIP) to compare the similarity between generated images and the text prompt. A higher CLIP score denotes higher similarity between prompt and generated image pairs. Besides CLIP, we introduce another alignment evaluation, an object detection evaluation that detects target objects using the Owl v2 \cite{minderer2024scaling} open-vocabulary object detection model and measures the detection success rate (denoted as Det.Rate). The detection success rate is the proportion of images that all target objects in the prompt can be successfully detected in the image. A higher detection success rate also implies a better alignment between prompts and generated images.
Moreover, we evaluate the perceptual similarity of the original image and the editing images through LPIPS Score (denoted as LPIPS) \cite{zhang2018perceptual}. A lower LPIPS score means fewer edits during the generation.
We also evaluate the generative quality of our image using Fréchet Inception Distance Score (denoted as FID) \cite{heusel2018gans}, which quantifies the discrepancy between the distribution of real images and that of the generated images.
Lastly, we evaluate the efficiency by measuring the average inference time of generating one image and report the computational overhead (denoted as Comp. Overhead) in percentage. The computational overhead is calculated as $\frac{T - T_0}{T_0}$, where $T$ is the inference time with edits and $T_0$ is the inference time of a clean diffusion model.

\begin{figure}[t]
    \centering
    \includegraphics[width=\textwidth]{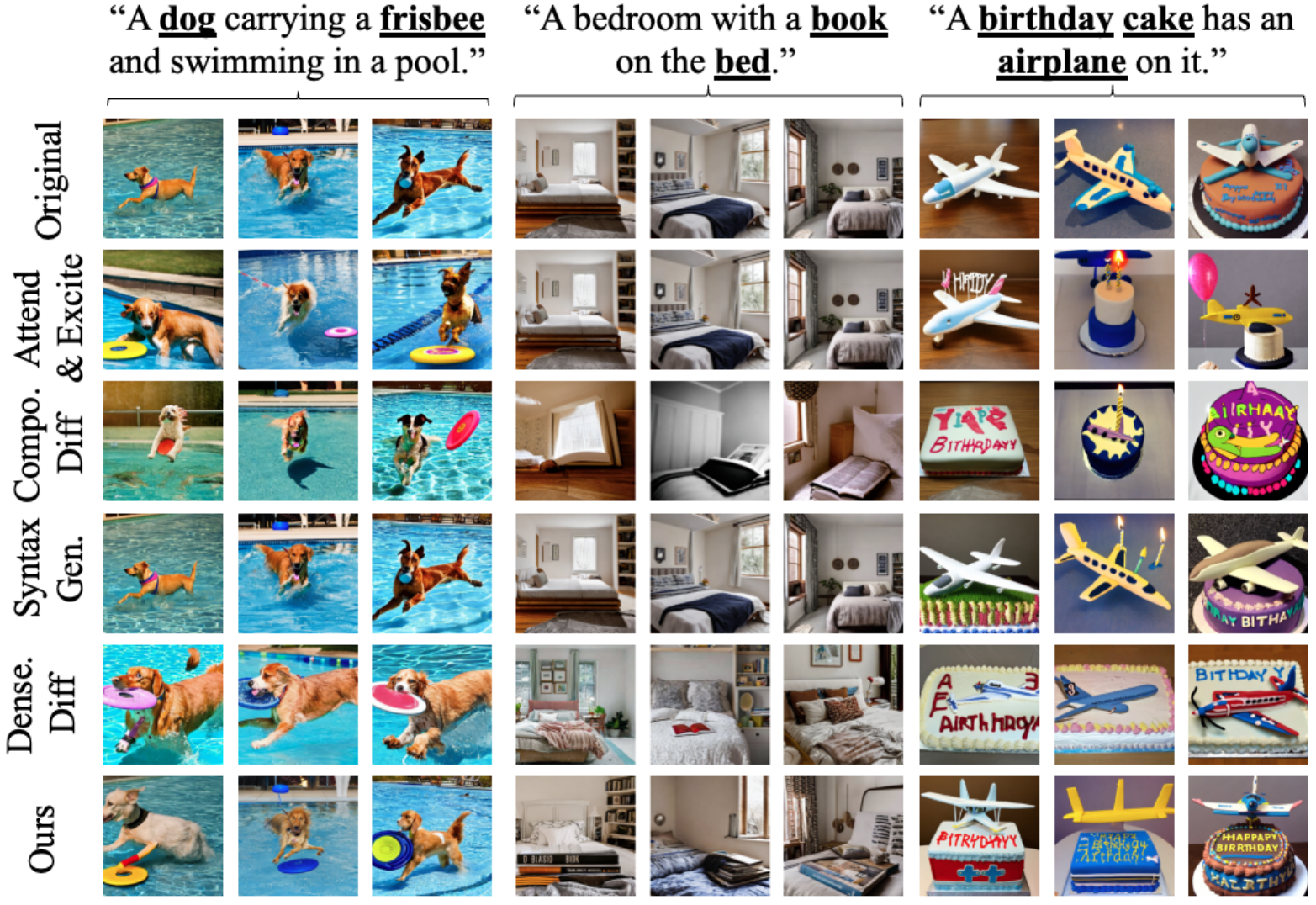}
    \caption{A qualitative comparison of the images generated by previous approaches and our approach. More samples in Appendix~\ref{appx:more_visual_comparison}.}
    \label{fig:visual_compare}
\end{figure}
\textbf{Datasets: }We include three datasets for our quantitative evaluation. One is a subset of the MS-COCO dataset (denoted as COCO Dataset) \cite{lin2014microsoft} proposed by the authors of \cite{kim2023dense}. This dataset slightly modifies captions from MS-COCO dataset with added attribute text for target words and uses modified captions as prompts. For this dataset, we can apply all baseline methods. The second dataset is a benchmark that analyzes semantic issues, created by the authors of Attend-And-Excite (denoted as A\&E Dataset) \cite{chefer2023attendandexcite}. For A\&E dataset, Dense Diffusion is not applicable, as it requires segmentation maps of targets as an additional input condition besides prompt texts. An accurate estimation of the inception distance for FID evaluation necessitates a substantial volume of data. Thus, we utilize a third distinct dataset, a subset of the MS COCO dataset comprising 3,000 images and 3,000 corresponding captions as prompts (denoted as FID Dataset).

\textbf{Diffusion models: }We evaluate on several diffusion models that perform text-to-image synthesis. Specifically, we include four open\-source diffusion models: Stable Diffusion $1.4$, $1.5$, $2$, and $2.1$. In our experiment, we generate $10$ images for each prompt with the default setting of each diffusion model.

\textbf{Hyperparameter search: }We perform a hyperparameter search to find the optimal hyperparameters of our method. Details and results of the hyperparameter search are in Section~\ref{sec:ablation_study}. For our method, we apply attention edit on the last down-sampling layer and the first up-sampling layer in U-Net (SD models have $3$ down-sampling layers and $3$ upsampling layers). We use $\beta = 0.1$ and stop edits after the $25^{th}$ diffusion step. 

\textbf{Hardware: } We perform T2I generation tasks on A4000 GPUs. The inference time for generating one image on SD models is around 4 seconds.

\subsection{Quantitative Comparison}
\begin{table}[t]
\centering
\caption{Evaluation of different methods on Stable Diffusion 2. The best results are shown in bold. DenseDiffusion is not applicable to the A\&E dataset due to the lack of segmentation masks. }
\label{tab:method_comparison}
\begin{tabular}{@{}c*{7}{c}@{}}
\toprule
{Methods} & \multicolumn{3}{c}{COCO Dataset} & \multicolumn{3}{c}{A\&E Dataset} & Comp. \\ 
\cmidrule(lr){2-4} \cmidrule(lr){5-7}
& CLIP$\uparrow$ & Det.Rate$\uparrow$ & LPIPS$\downarrow$  & CLIP$\uparrow$ & Det.Rate$\uparrow$ & LPIPS$\downarrow$  & Overhead$\downarrow$ \\
\midrule
Original & 0.328 & 55.9\% & --  & 0.324 & 46.7\% & -- & -- \\
\midrule
ComposableDiffusion & 0.301 & 27.0\% & 0.606  & 0.307 & 19.4\% & 0.598 & $119.5\%$ \\
\midrule
SyntaxGeneration & 0.327 & 63.8\% & 0.452  & 0.326 & 61.4\% & \textbf{0.423} & $378.0\%$ \\
\midrule
DenseDiffusion & 0.332 & 64.0\% & 0.729  & -- & -- & -- & $79.3\%$ \\
\midrule
AttendAndExcite & 0.330 & 64.5\% & \textbf{0.393}  & 0.331 & 60.8\% & 0.508 & $414.6\%$ \\
\midrule
Ours & \textbf{0.337} & \textbf{72.5\%} & 0.508  & \textbf{0.337} & \textbf{66.2\%} & 0.666 & \textbf{48.8\%} \\
\bottomrule
\end{tabular}
\end{table}
Figure~\ref{fig:visual_compare} presents sampled generation outcomes from various methods, utilizing Stable Diffusion 2 for generation. Figure~\ref{fig:visual_compare} demonstrates that attention regulation effectively enhances the semantic following ability of Stable Diffusion 2. 
To illustrate, in the left-most example of Figure~\ref{fig:visual_compare}, our method reliably produces all specified target objects (dog and frisbee), while other baseline methods fail to include all target objects.
We quantitatively measure the performance of all methods on two datasets, employing three metrics on prompt adherence, image quality in terms of generating target objects, and image similarity to images generated by the original model. Table~\ref{tab:method_comparison} shows quantitative evaluation results on Stable Diffusion 2. On both datasets, our method outperforms all baselines in terms of CLIP score and detection success rate. The evaluation results confirm that our attention regulation approach yields images that more accurately reflect the given prompts. Regarding computational efficiency, our method entails an additional 48\% computation time, markedly lower than the increase associated with other baseline methods. Furthermore, our approach introduces moderate adjustments, as indicated by an LPIPS score that is comparable to those of the baseline methods.

\begin{table}[t]
    \centering
    \caption{FID score evaluation. The best result is shown in bold. }
    \label{tab:fid}
    \begin{tabu}to \textwidth{X[c] X[c] X[c] X[c] X[c] X[c]}
    \toprule
    Methods & \centering Original & Compo.Diff. & SyntaxGen. & Attend\&Excite & Ours \\
    \midrule
    FID Score$\downarrow$ & 42.72 & 77.43 & \textbf{40.71} & 42.56 & 41.79 \\
    \bottomrule
    \end{tabu}
\end{table}
To evaluate the FID score of our attention regulation approach, we generate a single image for each prompt in our FID Dataset. Table~\ref{tab:fid} shows the FID score for our approach and other baselines. Our method achieves the second-lowest FID score, signifying superior generation quality relative to the other baselines.

We further explore the versatility of attention regulation across diverse diffusion models.
Table~\ref{tab:clip_diff_models} showcases additional experimental results, comparing our method with other baselines across different diffusion models. The CLIP score for our approach consistently outperforms those of the baselines, suggesting that attention regulation maintains its effectiveness across a variety of models. Due to the space constraints in the main text, we only report CLIP scores across multiple models. We present evaluation results of other metrics across multiple models in Appendix~\ref{appx:more_evaluation_metrics}.
\begin{table}[t]
\centering
\caption{CLIP score evaluation of different methods across diffusion models. The best results are shown in bold. We omit the result of other evaluation metrics for other diffusion models and present those results in \ref{appx:more_evaluation_metrics}.}
\label{tab:clip_diff_models}
\begin{tabular}{c c c c c c c c c}
\toprule
{Methods} & \multicolumn{4}{c}{COCO Dataset} & \multicolumn{4}{c}{A\&E Dataset} \\ 
\cmidrule(lr){2-5} \cmidrule(lr){6-9}
& SD1.4 & SD1.5 & SD2 & SD2.1 & SD1.4 & SD1.5 & SD2 & SD2.1 \\
\midrule
Original  & 0.323 & 0.324 & 0.328 & 0.323 & 0.314 & 0.314 & 0.324 & 0.327 \\
ComposableDiffusion & 0.306 & 0.305 & 0.301 & 0.305 & 0.308 & 0.308 & 0.307 &0.296\\
SyntaxGeneration & 0.325 & 0.325 & 0.327 & 0.324 & 0.323 & 0.323 & 0.326 & 0.326 \\
DenseDiffusion & 0.327 & 0.330 & 0.332 & 0.330 & -- & -- & -- & --\\
AttendAndExcite & 0.322 & 0.324 & 0.330 & 0.329 & 0.319 & 0.320 & 0.331 & 0.329 \\
Ours  & \textbf{0.331} & \textbf{0.332} & \textbf{0.337} & \textbf{0.335} & \textbf{0.331} & \textbf{0.331} & \textbf{0.337} & \textbf{0.337}\\
\bottomrule
\end{tabular}
\end{table}

\subsection{Ablation Study}\label{sec:ablation_study}
In this subsection, we delve into the impact of three critical hyperparameters on our attention regulation mechanism: the selection of attention layers, the timing of attention regulation within the diffusion steps, and the value of the $\beta$ regularization term.

\begin{figure}[t]
    \centering
    \begin{subfigure}{0.3\columnwidth}
        \includegraphics[width=\columnwidth]{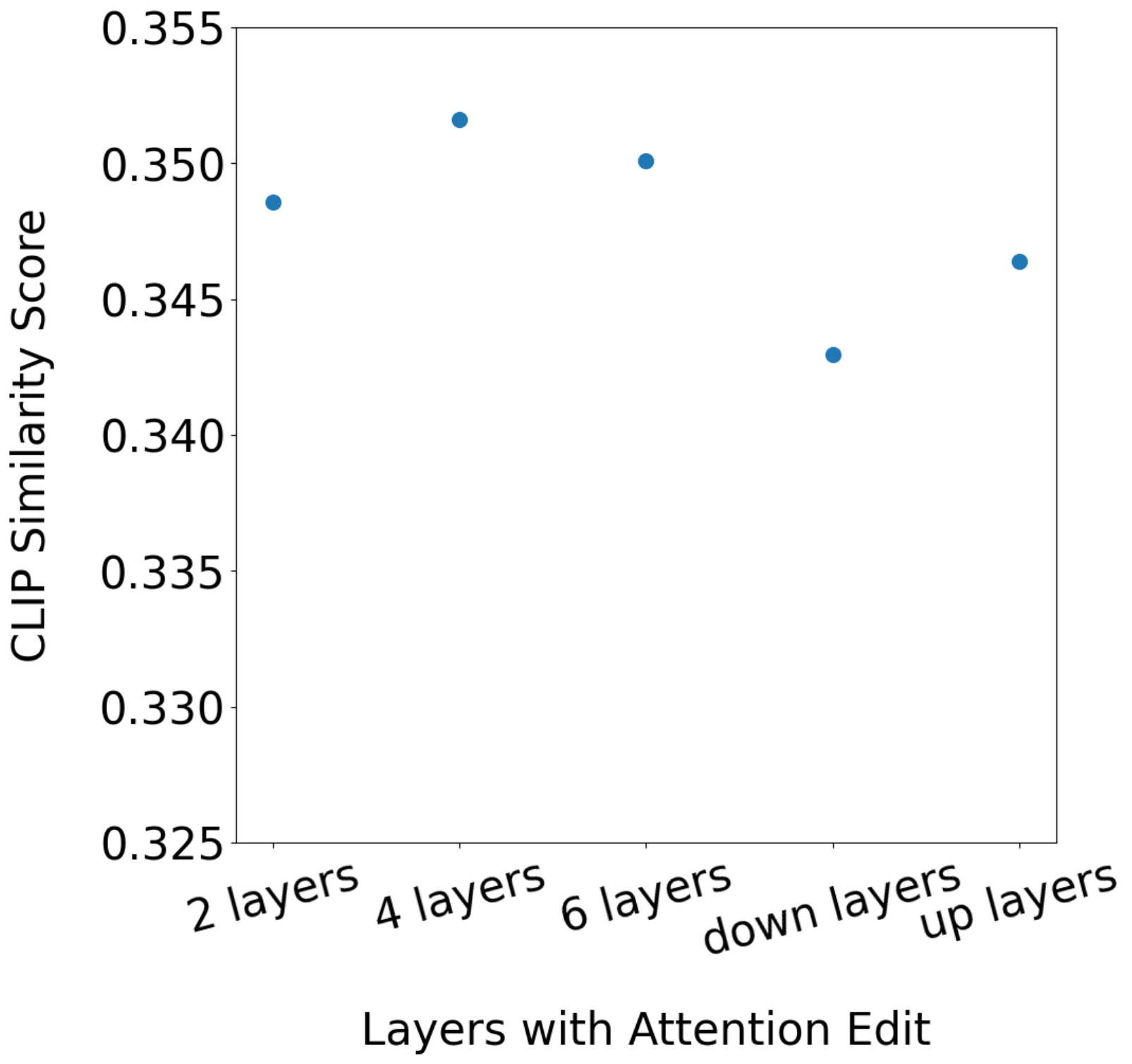}
        \caption{Layer Ablation}
        \label{fig:layer_ablation}
    \end{subfigure}
    \hfill 
    \begin{subfigure}{0.3\columnwidth}
        \includegraphics[width=\columnwidth]{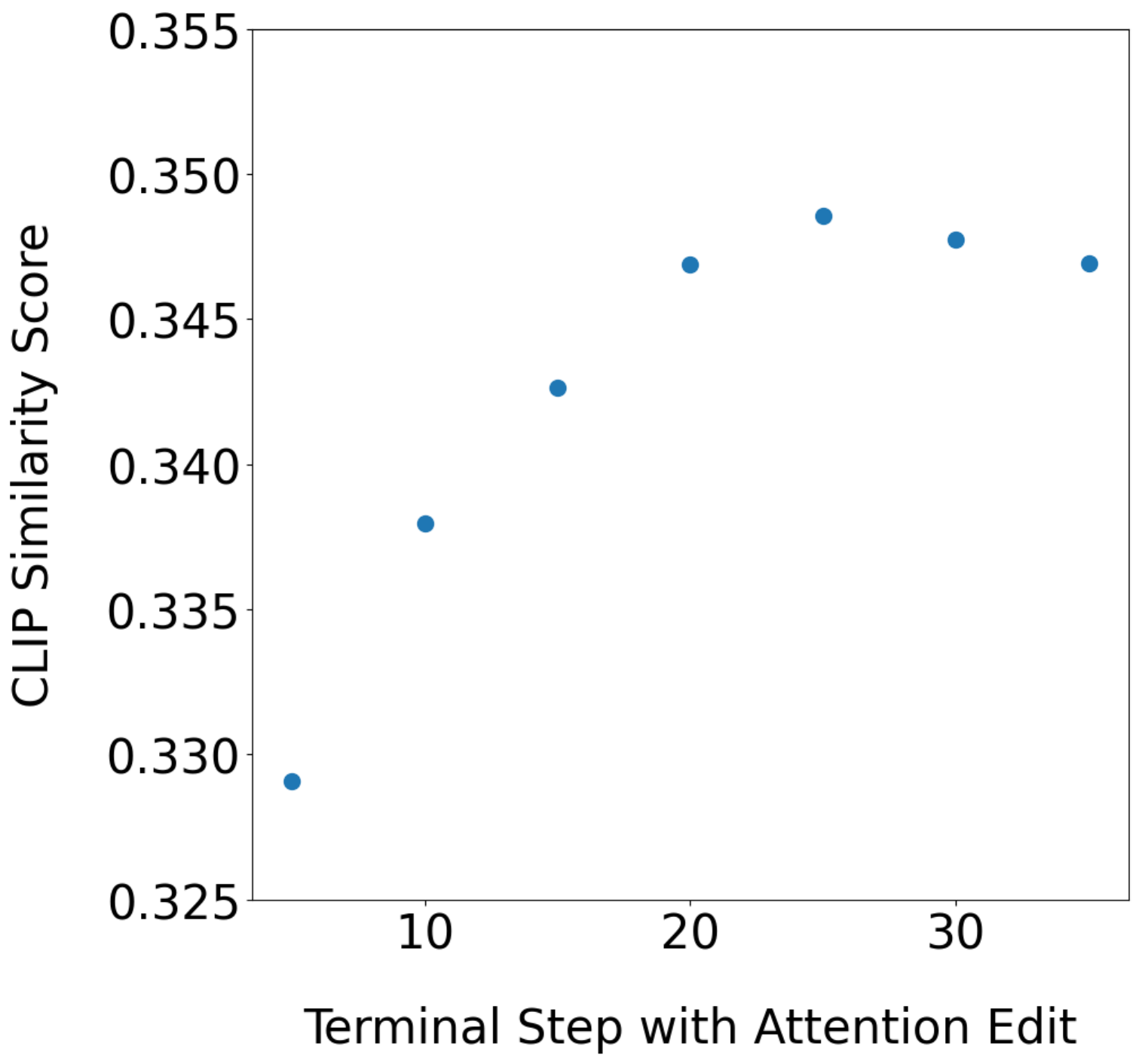}
        \caption{Step Ablation}
        \label{fig:step_ablation}
    \end{subfigure}
    \hfill 
    \begin{subfigure}{0.3\columnwidth}
        \includegraphics[width=\columnwidth]{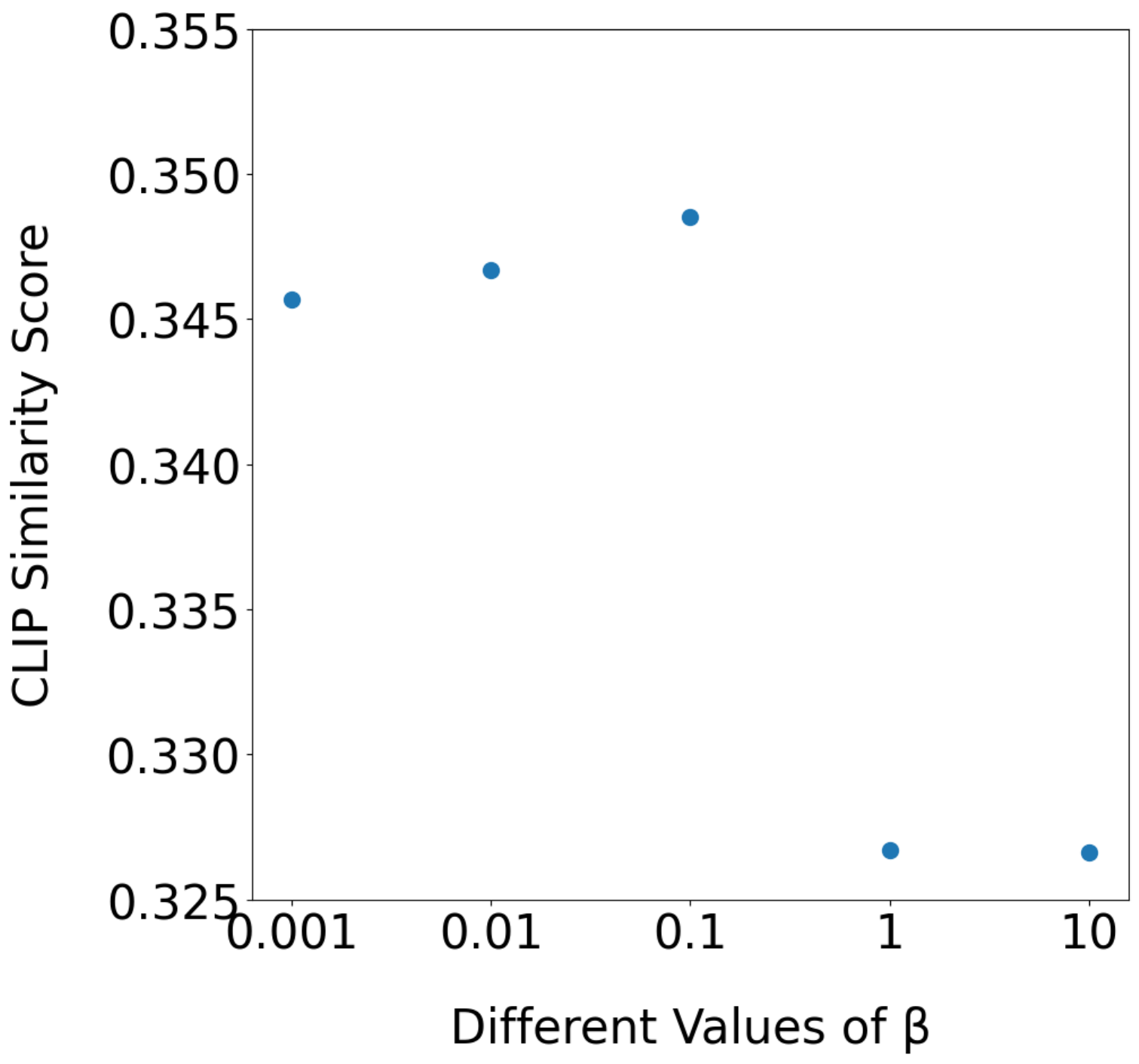}
        \caption{Beta Ablation}
        \label{fig:beta_ablation}
    \end{subfigure}
    \caption{Ablation study on layers (\ref{fig:layer_ablation}), diffusion steps (\ref{fig:step_ablation}) to perform attention regulation and the $\beta$ regularisation term (\ref{fig:beta_ablation}). Attention regulation performance increases initially by adding more layers and diffusion steps for editing, but saturates when reaching edit layer $4$ and edit steps $25$. The performance increases as $\beta$ increases until $0.1$.}
    \label{fig:ablation}
\end{figure}
\begin{figure}[t]
    \centering
    \includegraphics[width=0.9\textwidth]{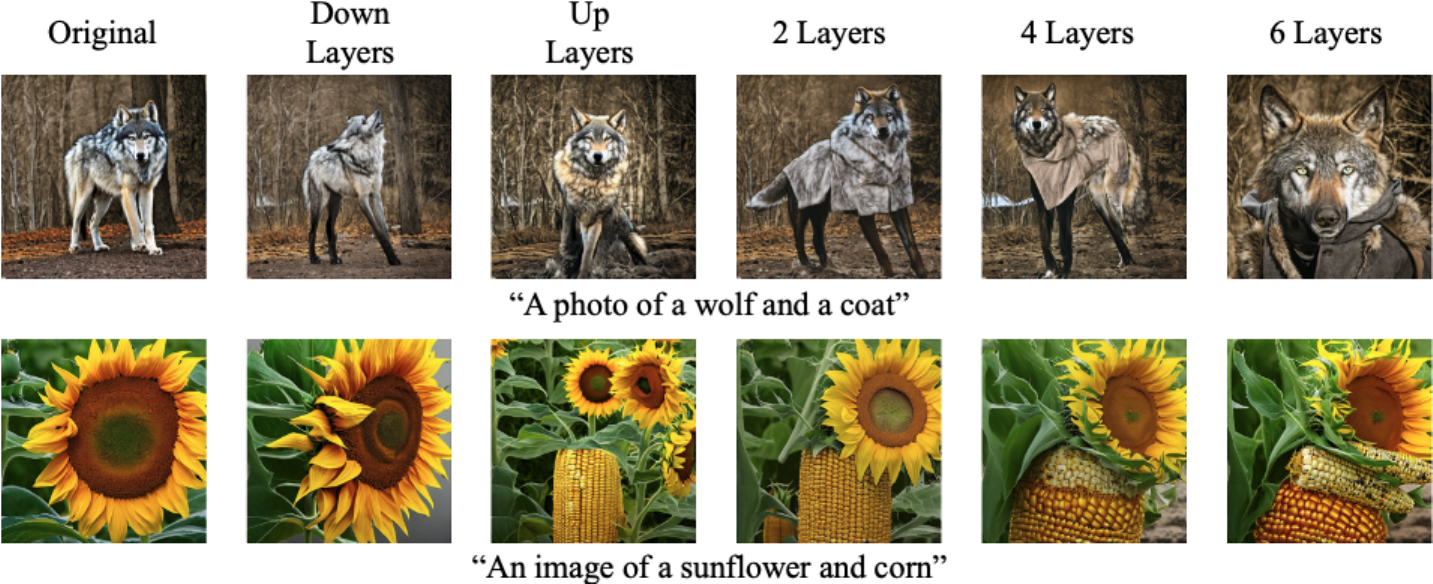}
    \caption{Example of attention regulation results with varying layers. The examples show that effective results can be achieved with just applying attention regulation on two cross-attention layers. More samples in Appendix~\ref{appx:more_ablation_results}.}
    \label{fig:ablation_layers_sample}
\end{figure}

Our analysis begins with the influence of integrating additional cross-attention layers. By methodically modifying more pairs of up and down block layers near the bottleneck layer in groups of one, two, and three, and assessing both down and up block layers individually, the best performance in CLIP similarity score is achieved with four layers, as illustrated in Figure~\ref{fig:layer_ablation}. However, to minimize complexity without significantly sacrificing performance, we opted for editing two layers, a decision supported by visual comparisons in Figure~\ref{fig:ablation_layers_sample}.

\begin{figure}[t]
    \centering
    \includegraphics[width=0.9\textwidth]{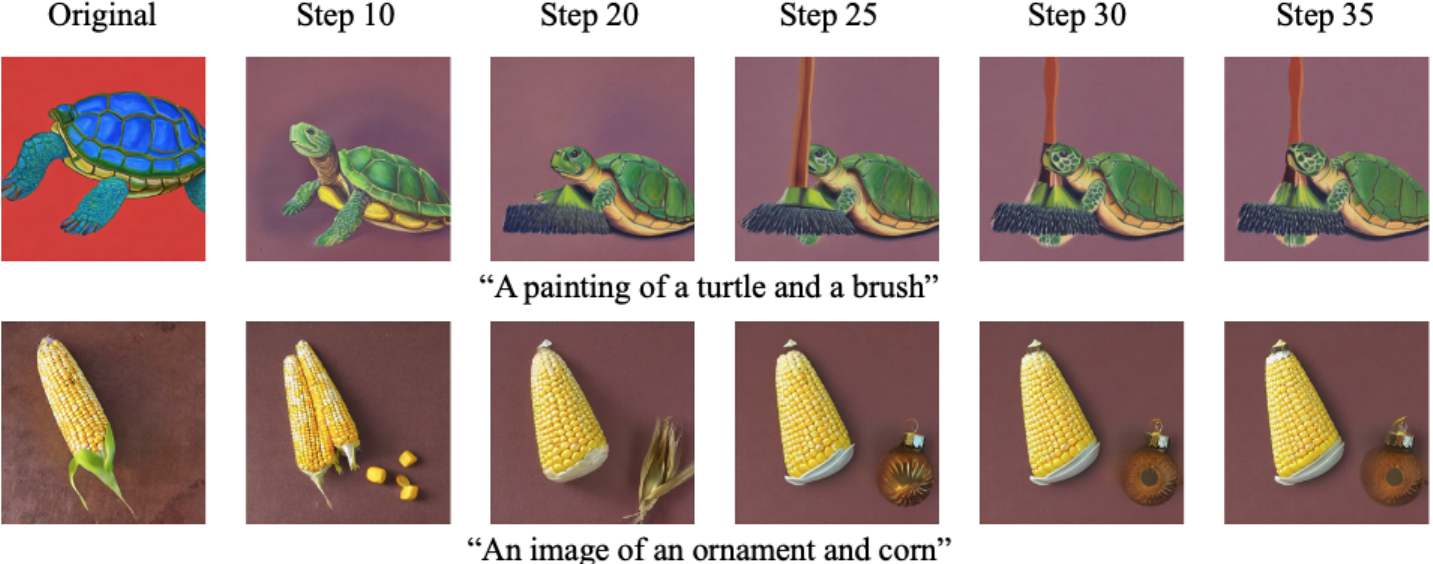}
    \caption{Example of attention regulation with varying steps. Visually, attention regulation is highly effective at step $25$. More samples in Appendix~\ref{appx:more_ablation_results}. }
    \label{fig:ablation_steps_sample}
\end{figure}

In addition, we explore how attention regulation at various diffusion steps affects outcomes by initiating regulation at the start and ceasing at different steps. Our observations, shown in Figure~\ref{fig:step_ablation}, reveal a performance plateau at step 25, after which the regulation's effectiveness decreases as the data becomes clearer. Figure~\ref{fig:ablation_steps_sample} illustrates the visual effects of stopping at different steps, leading to our decision to halt attention regulation at step 25.

Lastly, we adjust the $\beta$ regularization term to find a balance between edit efficacy and alterations to the original attention map. Increasing $\beta$ to $0.1$ enhanced performance marginally, but any higher value caused a significant decline, as depicted in Figure~\ref{fig:beta_ablation}. Thus, we establish $\beta$ at $0.1$ for optimal results. Visual examples to show the effects of varying $\beta$ are provided in \ref{appx:more_ablation_results}. 

\subsection{Limitations}\label{sec:limitation}
\begin{figure}[t]
    \centering
    \includegraphics[width=0.9\textwidth]{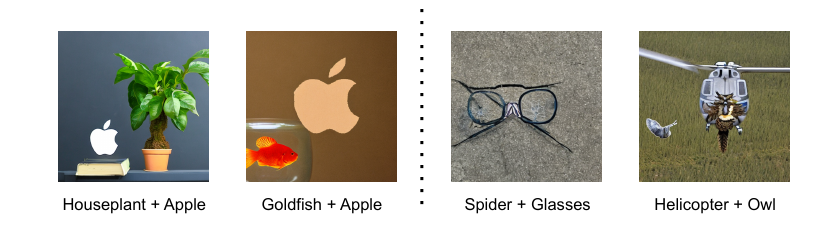}
    \caption{Failure cases of attention regulation. The left figures show the case of features not fit with the context. The right figures show the case of features fusion.}
    \label{fig:limitation}
\end{figure}

We demonstrate failure cases of our method and provide reasoning for the failure. Figure~\ref{fig:limitation} shows two types of failure. The two images on the left show that attention regulation may generate images distinct from human knowledge, as here the logo of Apple is incorrectly considered more appropriate by the model to improve semantic fidelity. For the two images on the right, SD2 fails to generate separate objects even with our attention regulation, instead, it fuses both concepts into one object. We conjecture both cases to be rooted in the features learned from diffusion models being different from human understanding. 
\section{Conclusion}
In this work, we study a prevalent issue in diffusion models equipped with cross-attention modules:  a single token often receives disproportionately high attention values during the generation process. To address the issue of dominating attention, we propose attention regulation, which modifies attention maps at inference time to enhance the semantic fidelity of generated images. We evaluate our method against baseline works across a wide spectrum and demonstrate the superior performance of our method in computation efficiency and adherence to prompts. Our method shows promise to be portable to existing diffusion models to enhance text conditioning without further modification on the model. 

\bibliographystyle{splncs04}
\bibliography{main}
\end{document}